\documentclass[9pt,twocolumn,twoside]{pnas-new}

\usepackage{subcaption}
\usepackage{multirow}

\graphicspath{ {images/} }

\templatetype{pnasresearcharticle} 

\title{Can neural networks predict dynamics they have never seen?}

\author[a,b,1]{Anton Pershin}
\author[b]{C\'edric Beaume} 
\author[b]{Kuan Li}
\author[b]{Steven M. Tobias}

\affil[a]{Atmospheric, Oceanic and Planetary Physics, University of Oxford, Oxford, UK}
\affil[b]{School of Mathematics, University of Leeds, Leeds, UK}

\leadauthor{Pershin} 

\significancestatement{Neural networks are a staple of 21st century technology, used in an increasingly wide range of applications, from automatic real-time speech recognition to early diagnosis of pathologies.
One of their stand-out successes is their skill in prediction of future dynamics given a suitable training set.
We show that Echo State Networks are able to predict  dynamical regimes that they have never encountered.
They succeed in predicting, both in the short and long term, transitions between two dynamical regimes, one of which it was not trained on. This paves the way for the use of neural network to explore new dynamics and provide early warnings of critical transitions typical for many physical, climate, biological and finance systems.}

\authorcontributions{A.P., C.B., K.L. and S.M.T. designed research; A.P. performed research; A.P., C.B., K.L. and S.M.T. analyzed
data; and A.P., C.B., K.L. and S.M.T. wrote the paper.}
\authordeclaration{The authors declare no conflict of interests.}
\correspondingauthor{\textsuperscript{1}To whom correspondence should be addressed. E-mail: anton.pershin@physics.ox.ac.uk}


\begin{abstract}
Neural networks have proven to be remarkably successful for a wide range of complicated tasks, from image recognition and object detection to speech recognition and machine translation.
One of their successes is the skill in prediction of future dynamics given a suitable training set of data. Previous studies have shown how Echo State Networks (ESNs), a subset of Recurrent Neural Networks, can successfully predict even chaotic systems for times longer than the Lyapunov time.
This study shows that, remarkably, ESNs can successfully predict dynamical behavior that is qualitatively different from any behavior contained in the training set.
Evidence is provided for a fluid dynamics problem where the flow can transition between laminar (ordered) and turbulent (disordered) regimes.
Despite being trained on the turbulent regime only, ESNs are found to predict laminar behavior.
Moreover, the statistics of turbulent-to-laminar and laminar-to-turbulent transitions are also predicted successfully, and the utility of ESNs in acting as an early-warning system for transition is discussed. 
These results are expected to be widely applicable to data-driven modelling of temporal behaviour in a range of physical, climate, biological, ecological and finance models characterized by the presence of tipping points and sudden transitions between several competing states.
\end{abstract}

\dates{This manuscript was compiled on \today}
\doi{\url{www.pnas.org/cgi/doi/10.1073/pnas.XXXXXXXXXX}}

\begin{document}

\maketitle
\thispagestyle{firststyle}
\ifthenelse{\boolean{shortarticle}}{\ifthenelse{\boolean{singlecolumn}}{\abscontentformatted}{\abscontent}}{}


\dropcap{N}eural networks are important examples of machine learning techniques that have shown tremendous image recognition, computer vision and speech recognition abilities.
Their utility stems from their ability to predict known behavior in new situations but how well they can extend this ability beyond their training set remains an open question \cite{Xu2021}.
An important aspect of this question is related to the prediction of previously unseen temporal behavior.
This becomes particularly interesting to explore given that neural networks have recently been introduced to assist physical modelling and, more generally, to time-dependent partial differential equations \cite{Brunton2020}, where the aim is to predict future dynamics without having to solve a computationally expensive set of equations.

Forecasting in dynamical systems is often achieved using a particular class of neural networks known as Recurrent Neural Networks (RNNs) \cite{Goodfellow2016}.
These are characterised by the presence of feedback connections within the network to allow it to ``remember'' the history of the dynamical system and use it to improve the accuracy of predictions.
Among the many different RNN architectures, we focus on Echo State Networks (ESNs) \cite{Jaeger2004, Lukovsevivcius2009} due to their relatively low training cost; compared with most other RNNs, for an ESN only one part of the network is trained while the rest is randomly generated and remains fixed \cite{Vlachas2020}.
Echo State Networks distinguished themselves by making sound short- and long-term predictions in various low-dimensional chaotic models \cite{Jaeger2004, Pathak2017, Chen2020, Doan2021}, in the Kuramoto--Sivashinsky equation \cite{Pathak2017, Pathak2018, Pathak2018modelfree} and in two-dimensional Rayleigh--B\'enard convection \cite{Pandey2020, Heyder2021}.
Trained with a single time series, ESNs can successfully approximate the statistical and geometric properties of chaotic attractors of a dynamical system \cite{Pathak2017, Haluszczynski2019, Chattopadhyay2020} and make short-term predictions on the level of accuracy corresponding to state-of-the-art techniques for time-series prediction while significantly outperforming them with respect to the memory and CPU usage \cite{Vlachas2020, Chattopadhyay2020}.

In this paper, we use ESNs to predict sudden transitions in fluids when a laminar (ordered) flow can undergo an instability and become turbulent (seemingly disordered) and vice-versa \cite{Barkley2016}.
Systems exhibiting qualitatively similar, bistable regimes are ubiquitous in both natural and engineering applications.
Examples include the transport of liquid and gases through pipelines, bioreactors in biochemical engineering, wind-turbines airfoils, as well as climate \cite{Lucarini2019}, ecological \cite{Menck2013}, Earth's magnetic field \cite{Petrelis2009} and  geodynamo \cite{Tobias2021} models.
Such transitions are often associated with a change in energy consumption or extreme damage which makes their prediction and, crucially, control an important task. 
We demonstrate that a properly trained ESN is capable of predicting both laminar-to-turbulent and turbulent-to-laminar transitions even if it has been trained using a time series containing only turbulent dynamics, i.e., it is able to infer laminar dynamics despite having only been trained on turbulent dynamics.

As an example of a transitional flow, we consider a paradigm model of plane Couette flow, i.e., the viscous flow between two parallel walls moving in opposite directions at constant and equal velocities.
The velocity field, $\boldsymbol{u}(\boldsymbol{x}, t)$ at position $\boldsymbol{x}$ and time $t$, is generally solved for via the integration of the Navier--Stokes equation accompanied by the incompressibility condition, no-slip boundary conditions in the wall-normal direction and spatial periodicity conditions in the streamwise $x$ and spanwise $z$ directions.
This set of equations can be reduced to the Moehlis--Faisst--Eckhardt (MFE) model \cite{Moehlis2004} by replacing plane Couette flow with a sinusoidal shear flow, known as the Waleffe flow \cite{Waleffe1997}, and proposing a truncation to nine Fourier-based modes, $\boldsymbol{u}_{j}(\boldsymbol{x})$, listed in SI Appendix, from which we can reconstruct the fluid velocity via:
\begin{equation}
    \boldsymbol{u}(\boldsymbol{x}, t) = \sum_{j = 1}^{9} a_j(t) \boldsymbol{u}_{j}(\boldsymbol{x}),
\end{equation}
where the time-dependent amplitudes are $\boldsymbol{a}(t) = [a_1(t), \dots, a_9(t)]$.
The nine-dimensional system of coupled amplitude equations obtained by projecting the Waleffe flow equations onto these modes reads
\begin{equation}
\label{eq:mhe}
    \frac{d}{dt} a_j = \delta_{1j} \frac{\pi^2}{4 Re} + \alpha_j(Re) a_j + \sum_{k=1}^{9} \sum_{l=1}^{9} \beta_{jkl}(Re) a_k a_l,
\end{equation}
where $Re$ is the Reynolds number, $\delta_{ij}$ is the Kronecker delta acting on indices $i$ and $j$, and $\alpha_j(Re)$ and $\beta_{jkl}(Re)$ are $Re$-dependent coefficients whose full expressions are given in SI Appendix.
The Reynolds number is the only physical parameter in this system.
It is a measure of the ratio between inertial and viscous forces.

\begin{figure*}
  \centering
  \includegraphics[width=0.95\linewidth]{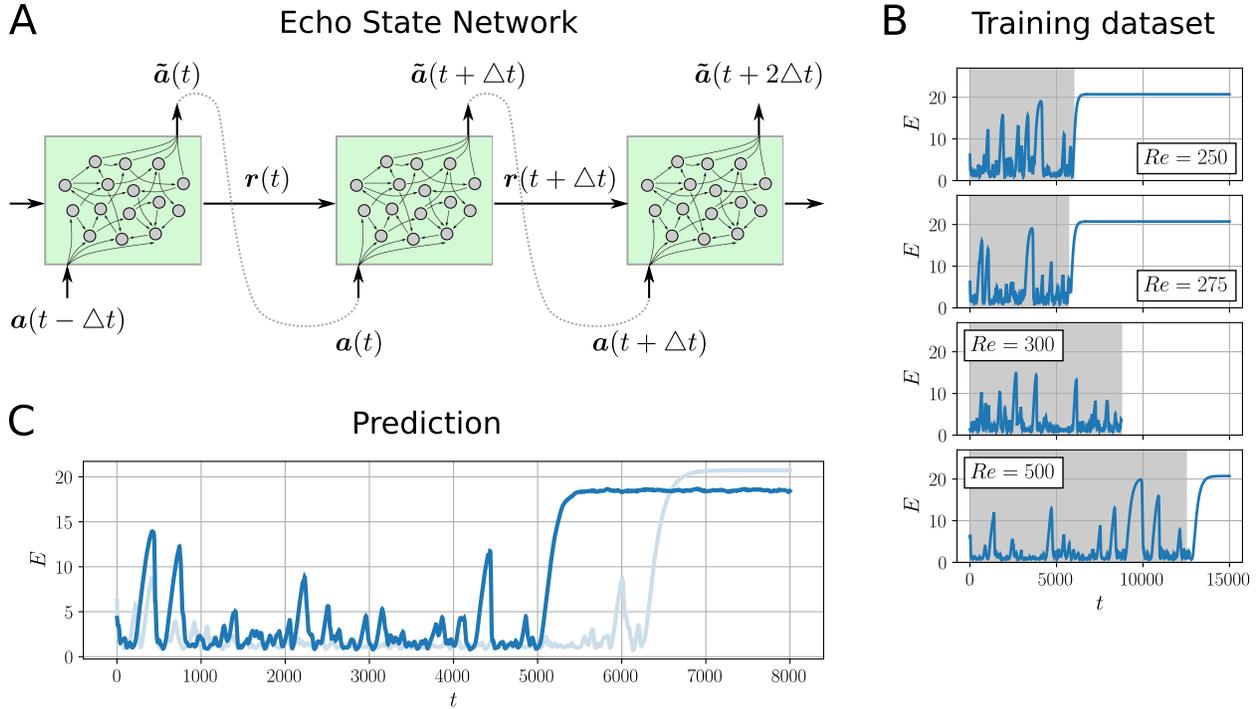}
  \caption{(top left) Schematic of an Echo State Network. In order to make a prediction $\boldsymbol{\tilde{u}}(t + \triangle t)$, the flow state $\boldsymbol{u}(t)$ at the previous time step and reservoir state $\boldsymbol{r}(t)$ are passed to the randomly generated reservoir (green box) where they are nonlinearly transformed to yield the prediction. (right) Training time series for Reynolds numbers $Re = 250$, $275$, $300$, $500$ obtained by time-integration of the Moehlis--Faisst--Eckhardt model and shown in the form of the time-evolution of the flow kinetic energy. Only shadowed parts were used for training. (bottom) Flow prediction made by the Echo State Network trained at $Re = 300$ (bright blue curve) and a representative turbulent trajectory of the Moehlis--Faisst--Eckhardt model computed at the same Reynolds number (light blue curve).}
  \label{fig:main_panel}%
\end{figure*}

The only known stable solution of \eqref{eq:mhe} is the steady laminar flow: $\boldsymbol{a}_{\text{lam}} = [1, 0, \dots, 0]^T$ which is equivalent to $\boldsymbol{u}_{\text{lam}} = \sqrt{2} \sin(\pi y / 2) \boldsymbol{e}_x$ in  physical space; here $\boldsymbol{e}_x$ is the unit vector in the $x$-direction.
Despite the stability of the laminar flow, we can observe long-lived turbulence for $Re \gtrsim 150$ \cite{Moehlis2004}.
Examples of turbulent flows at different values of the Reynolds numbers are shown in figure \ref{fig:main_panel}B through timeseries of the kinetic energy:
\begin{equation}
    E = \frac{1}{2} ||\boldsymbol{u}||_2^2 = \Gamma_x \Gamma_z \sum_{j=1}^{9} a_j^2,
\end{equation}
where $\Gamma_x = 1.75 \pi$ (resp. $\Gamma_z = 1.2$) is the imposed solution wavelength in the $x$ (resp. $z$) direction. 
All our simulations display long-lasting chaotic dynamics but eventually relax to the laminar flow, which is expected to be the global attractor at least for $Re \lesssim 335$ \cite{Moehlis2005}.
This phenomenon, called hereafter turbulent-to-laminar transition, is a prominent feature of transitional shear flows \cite{Barkley2016}.
The opposite process of laminar-to-turbulent transition is equally important both from a theoretical and a practical viewpoint.
In this study, we show that statistical features associated with both laminar-to-turbulent and turbulent-to-laminar transitions can be successfully predicted by an ESN trained solely on a transient segment of a turbulent trajectory, i.e., with no experience of laminarization.

\section*{\label{sec:rc}Method}

Echo State Networks belong to a class of artificial Recurrent Neural Networks (RNNs) that are characterized by the presence of internal feedback connections in their architectures allowing a network to have its own ``memory'' and, thereby, generate time series with a greater accuracy in comparison to its non-recurrent companions.
Figure \ref{fig:main_panel}A shows a schematic representation of a typical RNN architecture that takes the flow state $\boldsymbol{a}(t) \in \mathbb{R}^{N_a}$ at time $t$ as an input, where $N_a = 9$ for the Moehlis--Faisst--Eckhardt model, and outputs the prediction $\boldsymbol{\tilde{a}}(t + \triangle t)$ of the flow state at time $t + \triangle t$, where $\triangle t = 1$ throughout this study.
In addition to the input flow state, the RNN uses its own internal state $\boldsymbol{r}(t) \in \mathbb{R}^{N_r}$ which is passed through the feedback connection.
In the context of ESNs, $\boldsymbol{r}(t)$ is called the reservoir state.
Our ESN uses a two-step prediction.
First, the reservoir state $\boldsymbol{r}(t)$ and the input flow state $\boldsymbol{a}(t)$ are nonlinearly transformed to get the reservoir state at time $t + \triangle t$:
\begin{equation}
\label{eq:esn_reservoir_state}
    \boldsymbol{r}(t + \triangle t) = \tanh \left[\boldsymbol{b} + \boldsymbol{W}\boldsymbol{r}(t) + \boldsymbol{W}_{in}\boldsymbol{a}(t)\right] + \xi \boldsymbol{Z},
\end{equation}
where $\boldsymbol{W}$ and $\boldsymbol{W}_{in}$ are $N_r \times N_r$ and $N_r \times N_a$ weight matrices, $\boldsymbol{b}$ is a $N_r$-dimensional bias vector, $\boldsymbol{Z}$ is a random vector uniformly distributed between $-0.5$ and $0.5$ and $\xi$ is a hyperparameter controlling the amplitude of the additive noise.
Second, the reservoir state is mapped back into the flow space via the linear transformation:
\begin{equation}
\label{eq:esn_output_state}
    \boldsymbol{\tilde{a}}(t + \triangle t) = \boldsymbol{W}_{out}\begin{bmatrix}\boldsymbol{r}(t + \triangle t) \\ 1\end{bmatrix},
\end{equation}
where $\boldsymbol{W}_{out}$ is a $N_a \times (N_r + 1)$ weight matrix.
The result, $\boldsymbol{\tilde{a}}(t + \triangle t)$, is called the prediction.
The network is trained so that the prediction approximates the true flow state $\boldsymbol{a}(t + \triangle t)$ as accurately as possible.

The characteristics that distinguishes ESNs from the vast majority of other RNN architectures is that the weight matrices $\boldsymbol{W}$ and $\boldsymbol{W}_{in}$ and bias term $\boldsymbol{b}$ are initialized randomly and remain fixed, i.e., they are not trained, and the weight matrices are often chosen to be sparse which makes the network sparsely connected (see SI Appendix for details).
This greatly simplifies the training process which becomes equivalent to solving the linear regression problem:
\begin{equation}
    \label{eq:min_rss}
    \min_{\boldsymbol{W}_{out}} \sum_{k=1}^{N_t} || \boldsymbol{W}_{out} \boldsymbol{r}(k \triangle t) - \boldsymbol{a}(k \triangle t) ||_2^2,
\end{equation}
where it is assumed that the training dataset is composed of $N_t+1$ flow states $\boldsymbol{a}(t)$ known at times $t = 0, \triangle t, 2\triangle t, \dots, N_t\triangle t$.
The flow state at $t = 0$ is used as an initial condition only to compute the first prediction $\tilde{\boldsymbol{a}}(\triangle t)$.
This minimization problem possesses a closed-form solution given by the normal equation:
\begin{equation}
    \boldsymbol{W}_{out}^T = \left( \boldsymbol{R}^T \boldsymbol{R} \right)^{-1} \boldsymbol{R}^T \boldsymbol{A},
\end{equation}
where matrix $\boldsymbol{R} \in \mathbb{R}^{N_{t} \times (N_{r} + 1)}$ is made of vectors $\boldsymbol{r}(\triangle t), \boldsymbol{r}(2\triangle t), \dots, \boldsymbol{r}(N_t \triangle t)$ and an all-ones vector and $\boldsymbol{A} \in \mathbb{R}^{N_{t} \times N_{a}}$ is made of vectors $\boldsymbol{a}(\triangle t), \boldsymbol{a}(2\triangle t), \dots, \boldsymbol{a}(N_t \triangle t)$.
We wish to emphasize two modifications which differ our ESN architecture from more standard alternatives found in the literature.
The first one is the random bias term $\boldsymbol{b}$ in equation \eqref{eq:esn_reservoir_state}. This significantly improves the accuracy of predictions in our case.
The second one is the additive noise in the same equation which is introduced to regularize the regression problem and, at the same time, improve the stability of our ESN \cite{Lukovsevivcius2009}.

The aforementioned architecture involves several hyperparameters: the reservoir state dimension $N_r$, the spectral radius $\rho(\boldsymbol{W})$ of matrix $\boldsymbol{W}$, its sparsity $s$ and the noise amplitude $\xi$.
Though we did not perform an exhaustive search of optimal hyperparameter values, several points need to be highlighted.
The success of ESNs relies on a high-dimensional reservoir space whose dimension $N_r$ is expected to be much higher than that of the flow state.
Consequently, we fix the reservoir state dimension at a relatively large value $N_r = 1500$.
We also fix the noise amplitude $\xi = 10^{-3}$ and spectral radius $\rho(\boldsymbol{W}) = 0.5$ which was observed to minimize the residual sum of squares in \eqref{eq:min_rss}.
Surprisingly, we found a weak dependence of the sparsity on the quality of prediction and used the value of $s = 0.5$ for $Re = 200$ and $s = 0.9$ for other Reynolds numbers.
See SI Appendix for further details of the hyperparameter search.

Once our ESN is built, we can use it to make predictions of the flow state.
For this, we simply replace $\boldsymbol{a}$ in \eqref{eq:esn_reservoir_state} with $\boldsymbol{\tilde{a}}$ which is equivalent to activating one more feedback connection (gray dotted line in figure \ref{fig:main_panel}A).
This makes \eqref{eq:esn_reservoir_state} and \eqref{eq:esn_output_state} a closed system of recurrent equations which only requires initial conditions $\boldsymbol{\tilde{a}}(0)$ and $\boldsymbol{r}(0)$.
However, the initial reservoir state $\boldsymbol{r}(0)$ is not known in advance and must be found through a so called synchronization process.
Namely, we take a small number of states from the recent flow history: $\boldsymbol{a}(-9 \triangle t), \boldsymbol{a}(-8 \triangle t), \dots, \boldsymbol{a}(-\triangle t)$ and subsequently generate reservoir states $\boldsymbol{r}(-8 \triangle t), \boldsymbol{r}(-7 \triangle t), \dots, \boldsymbol{r}(0)$ using equation \eqref{eq:esn_reservoir_state} and zero initial condition $\boldsymbol{r}(-9 \triangle t) = \boldsymbol{0}$.
At the end of this synchronization process, we obtain the required initial reservoir state $\boldsymbol{r}(0)$ and can use it to predict the flow dynamics following the procedure described above.

\section*{\label{sec:results}Results}

We provide evidence that ESNs are able to predict laminar dynamics without having observed it before.
We first generate transient turbulent trajectories by time-integrating random initial conditions using a fourth-order Runge--Kutta scheme applied onto \eqref{eq:mhe} with time step $10^{-3}$.
One such trajectory is generated for each of the following Reynolds numbers: $Re = 250, 275, 300$ and $500$ (see figure \ref{fig:main_panel}B).
As these simulations eventually relax to the laminar flow ($E \approx 20.7$), we selected the training set to only comprise turbulent dynamics, as shown by the shaded regions in figure \ref{fig:main_panel}B.
We then train one ESN per Reynolds number using the resulting training sets.
To ease the notation, hereafter we label the results produced by the Moehlis--Faisst--Eckhardt model as  the ``truth'' and those of the associated ESN as the ``prediction''.

Each of the trained ESNs is able to generate turbulent trajectories whose statistical properties are similar to that of the original model.
This fact has already been established in \cite{Doan2021} where ESNs are used to predict statistical properties and extreme events associated with the dynamics of the Moehlis--Faisst--Eckhardt model.
In this paper, we demonstrate that ESNs are able to handle a more non-trivial task of predicting laminarization, i.e., the decay process of a turbulent trajectory towards a laminar state it has not encountered before, as exemplified in figure \ref{fig:main_panel}C.
At $t \approx 5000$, the predicted flow (dark blue curve) terminates its low-energy chaotic oscillations to relax to a higher energy behavior with only weak temporal variations attributed to the presence of small-amplitude noise in equation \eqref{eq:esn_reservoir_state}.
This new behavior is akin to the true laminar state, located at $E \approx 20.7$ 

Despite the difference between the true and the predicted laminar flow, it is particularly noteworthy that the ESN is able to predict laminarization, a transition to which it was not exposed during training.
We note that qualitatively the same prediction was obtained even if we turned the noise off in equation \eqref{eq:esn_reservoir_state} while making prediction.
In contrast to figure \ref{fig:main_panel}C, the predicted laminar state in this case would be an equilibrium without any weak variations.
However, we found that the presence of noise leads to more accurate representation of transition statistics which motivated us to use it throughout the study.
It is also important to emphasize that the transition occurring in our prediction at $t \approx 5000$ is different from the ``amplitude death'' phenomenon, recently investigated in \cite{Xiao2021}: our main parameter $Re$ is not changed dynamically.

As we shall see in the next sections, ESNs are capable of more surprising predictions, such as transition statistics.
First, we demonstrate their skill in learning turbulent-to-laminar transition by showing that ESNs can successfully recover the distribution of lifetimes of turbulent trajectories \cite{Avila2011}.
Additionally, we provide evidence of their ability to make short-term probabilistic predictions of transitional events. This paves the way for their use as generators of early-warning signals of critical transitions \cite{Scheffer2009}.
Finally, we take a look at the opposite kind of transition, laminar-to-turbulent transition, and show that ESNs can be used to approximate the transition probability, one of the key statistics associated with this type of instability \cite{Pershin2020}.


\subsection{\label{subsec:lifetime}Turbulent-to-laminar transition}

Turbulent-to-laminar transitions are often characterized using statistical tools similar to the survival function $S(t) = P(T \geq t)$, which represents the probability that turbulent behavior remains observed for a duration $t$ or, equivalently, that the time $T$ at which the laminarization event eventually takes place is larger than $t$ \cite{Moehlis2004, Avila2011, Barkley2016}.
Within the context of the Moehlis--Faisst--Eckhardt model, for $Re \lesssim 300$, this distribution takes the form \cite{Moehlis2004}:
\begin{equation}
\label{eq:exponentiallaw}
    S(t; Re) = \exp\left[\frac{t - t_0}{\tau(Re)}\right],
\end{equation}
where $t_0$ is the time delay caused by the approach of the turbulent saddle and $1/\tau(Re)$ is the $Re$-dependent escape rate.
To build the lifetime distribution for the original MFE model at a fixed value of the Reynolds number, we time-integrate $200$ random initial conditions generated by drawing initial amplitudes $a_j(0)$ from the uniform distribution with support $[-1; 1]$ such that
the kinetic energy of any initial condition is equal to $E = 0.3 \Gamma_x \Gamma_z$ as described in \cite{Moehlis2004}.
The lifetime $T$ is measured for each of these initial conditions in the following way: we assume that a laminarization event has taken place and, thus, record $T$, if the total kinetic energy of the flow $E > 15$ from time $T - 1000$ at least until time $T$.

\begin{figure}
    \includegraphics[width=0.49\textwidth]{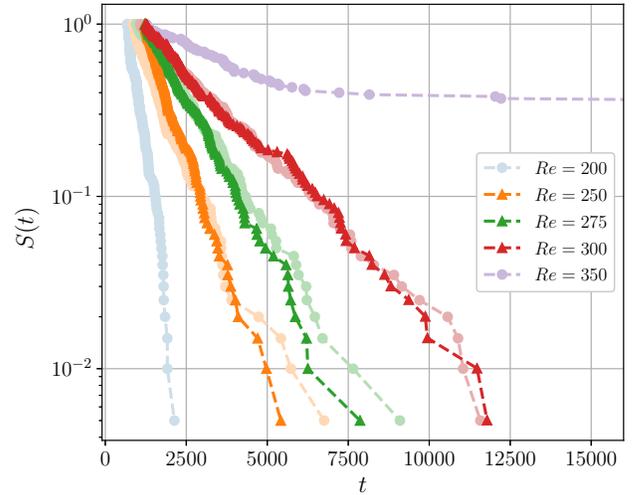}
    \caption{Lifetime distributions for Reynolds numbers from $Re = 200$ to $Re = 350$ shown in the form of survival functions. Bright (resp. light) colours correspond to the distributions generated by the Echo State Networks (resp. Moehlis--Faisst--Eckhardt model).}
    \label{fig:lifetime}%
\end{figure}

The above procedure is used for $Re = 200, 250, 275, 300, 350$ and the resulting survival functions are shown in figure \ref{fig:lifetime} (light colours).
For $Re = 350$ and beyond, the lifetime distribution does not follow law \eqref{eq:exponentiallaw}, as was already observed in \cite{Moehlis2004}, so we did not investigate such values of the Reynolds number.
The same procedure is repeated using our trained ESNs to provide the dark color curves in figure \ref{fig:lifetime}.
We did not obtain results for $Re = 200$ owing to the fact that laminarization occurs too soon to generate a sufficiently long laminarization-free time-series for training.
The predictions of our ESNs are excellent.
The predicted lifetime distributions preserve the main qualitative feature of the true distributions, their exponential structure, implying that the memoryless nature of the laminarization process has been adequately learned.
Furthermore, the escape rate of these survival laws, $1/\tau(Re)$, is also well-predicted as one can observe in table \ref{tbl:mle_exp_law}, which is a surprising result, given that our ESNs were not trained on data sets including laminarization events.

\begin{table}\centering
\caption{Maximum Likelihood Estimates of parameters $t_0$ and $\tau(Re)$ of the exponential distribution \eqref{eq:exponentiallaw} approximating lifetime distributions computed both for the Moehlis--Faisst--Eckhardt model and ESNs.}
\label{tbl:mle_exp_law}
\begin{tabular}{lccccc}
\multirow{2}{*}{$Re$} & \multicolumn{2}{c}{$t_0$} & \multicolumn{2}{c}{$\tau(Re)$} & \multirow{2}{*}{Relative error in $\tau(Re)$} \\
& Truth & Prediction & Truth & Prediction & \\
\midrule
250 & 845 & 1207 & 835 & 734 & 0.121 \\
275 & 956 & 1222 & 1202 & 1249 & 0.169 \\
300 & 1086 & 1248 & 2161 & 2089 & 0.033 \\
\bottomrule
\end{tabular}
\end{table}

\subsection{\label{sec:short_term_transition}Early warning of turbulent-to-laminar transition}

Lifetime distributions, such as those considered above, are used to predict statistically  the long-term behavior of turbulent trajectories.
In many cases, however, it is a short-term prediction that is of interest. It can be used to anticipate critical transitions in, for example, climate \cite{Lenton2011, Ashwin2012}, geophysical \cite{Ditlevsen2010}, ecological \cite{Kefi2014} and many other complex nonlinear systems \cite{Scheffer2009, Scheffer2012}.

In the transitional flow problem considered here, we may want to determine whether a given turbulent flow $\boldsymbol{u}(t)$ will laminarize within a relatively short time window, e.g. $T = 2000$.
In the case of a deterministic system, such as the Moehlis--Faisst--Eckhardt model, it is sufficient to time-integrate a given initial condition to learn whether the laminarization event occurs.
Our ESN, in contrast, is a stochastic model by design owing to the presence of the noise term in \eqref{eq:esn_reservoir_state} and, thus, can rather be used to assess the probability of turbulent-to-laminar transition within a given time window.
To compute this probability, we perform ensemble predictions where each prediction within the ensemble starts from the same initial condition but, due to the noise, evolves differently from other predictions.
The probability of turbulent-to-laminar transition is then computed as the fraction of these predictions leading to laminarization.


We start from exploring the average level of ``leakiness'' of the turbulent saddle generated by the ESN at $Re = 500$, i.e. the average probability that a typical turbulent flow suddenly laminarizes given a short sequence of its previous states.
This value acts as a reference for future predictions.
To make such a measurement with respect to the ESN, we pick $N = 100$ random states $\boldsymbol{a}(t_j)$, $j = 1, \dots, N$, from the test dataset. This dataset was not used for training nor search for the hyperparameter values, and make ensemble predictions for each of them assuming that the synchronization is done with respect to a sequence of $10$ initial states $\boldsymbol{a}(t_j - 9\triangle t), \boldsymbol{a}(t_j - 8\triangle t), \dots, \boldsymbol{a}(t_j)$.
Each ensemble member corresponds to a prediction for the first $2000$ time units and is classified as either exhibiting turbulent-to-laminar transition or not.
The probability of turbulent-to-laminar transition, denoted as $P_{\text{T}\to \text{L}}(t_j)$, is then estimated as a fraction of laminarizing trajectories. 
The average probability of turbulent-to-laminar transition, which we will refer to as the reference probability $P_{\text{ref}}$, is obtained by averaging $P_{\text{T}\to \text{L}}(t_j)$ with respect to $t_j$ and appears to be equal to $P_{\text{ref}} \approx 0.11$.
\begin{figure}
\includegraphics[width=0.49\textwidth]{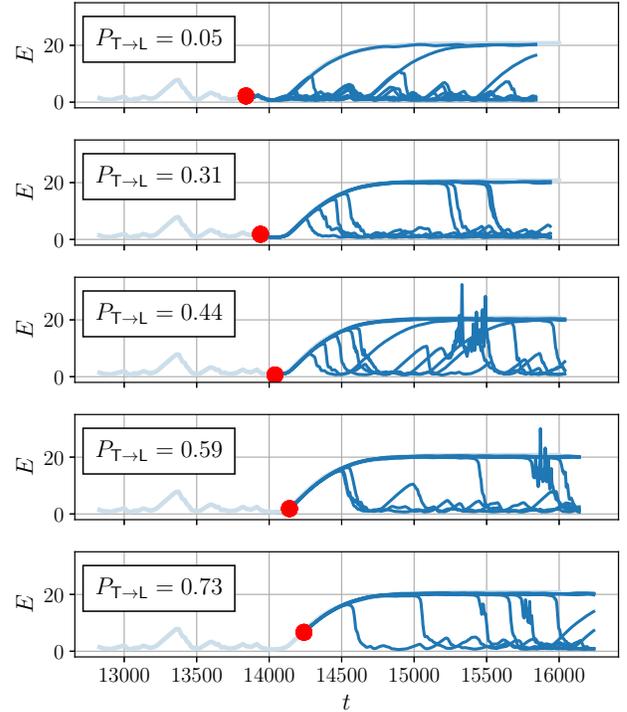}
\caption{Prediction of turbulent-to-laminar transition at $Re = 500$ based on the calculation of the probability of turbulent-to-laminar transition $P_{\text{T}\to \text{L}}$ which is estimated using an ensemble approach for four initial states (red dots) at times $t_j = 13840 + 100(j - 1)$, where $j = 1, \dots, 5$. Every $10$th ensemble member of the prediction generated by the Echo State Network is plotted in bright blue. The true flow trajectory obtained by time-integration of the Moehlis--Faisst--Eckhardt model is plotted in light blue.
}
\label{fig:prediction_of_turbulent_to_laminar_transition}
\end{figure}
Having this value, we can formulate a rule for the early warning of turbulent-to-laminar transition: if the probability $P_{\text{T}\to \text{L}}(t)$ computed at time of observation $t$ is significantly larger than $P_{\text{ref}}$, then we expect a turbulent flow to relax to the laminar state in a relatively short amount of time after $t$.

We demonstrate that our ESNs are able to act as generators of early warning signals by making probabilistic predictions of a transition starting approximately at $t \approx 14000$ for the turbulent flow at $Re = 500$ shown in figure \ref{fig:main_panel}B.
Since a non-shadowed piece covering this transition was not used for training purposes, we can safely use it.
We expect that the probabilities of turbulent-to-laminar transition $P_{\text{T}\to \text{L}}(t)$ estimated by the ESN become higher as $t$ approaches the point of the beginning of laminar dynamics.
To verify this, we pick four initial states at times $t_j = 13840 + 100(j - 1)$, where $j = 1, \dots, 5$, and compute the corresponding probabilities of turbulent-to-laminar transition $P_{\text{T}\to \text{L}}(t_j)$ using the ESN trained at $Re = 500$ and $N = 100$ ensemble members for each initial state.
As required by the synchronization procedure, we also use $9$ flow states prior to each given initial state.
The probability of turbulent-to-laminar transition is then computed using exactly the same ensemble-based algorithm as we used to compute the reference probability $P_{\text{ref}}$.
The resulting probabilities together with a small selection of ensemble members used for the estimation are shown in figure \ref{fig:prediction_of_turbulent_to_laminar_transition}.
We note a clear trend of the probability $P_{\text{T}\to \text{L}}(t_j)$ increasing as the initial time $t_j$ approaches that of the laminarization event which is consistent with our expectations.
It starts at $t_0 = 13840$ from the probability $0.05$ (top panel in figure \ref{fig:prediction_of_turbulent_to_laminar_transition}), a value comparable to the reference value $P_{\text{ref}} \approx 0.11$ and thereby not implying any likely transition to the laminar state.
At slightly later initial time $t_1 = 13940$, the probability of turbulent-to-laminar transition jumps up to a larger-than-reference value $0.31$, which can already be considered as early warning, and further grows steadily as we increase the initial time. 
At the latest stage, the probability of turbulent-to-laminar transition reaches the value of $0.73$ implying that the ESN considers the turbulent flow relaxation to the laminar state in a short time as a very likely event.
It is important to emphasize that all the probabilities for $t \geq 13940$ are substantially larger than $P_{\text{ref}} \approx 0.11$ thereby confirming that they indeed act as early warning signals.

\subsection{\label{subsec:p_lam}Laminar-to-turbulent transition}


The transition from turbulence to laminar flow does not follow similar dynamical processes to its reciprocal laminar-to-turbulent transition.
While the former is a sudden escape from a turbulent saddle (i.e., not an attractor), transition to turbulence is a finite-amplitude instability: the laminar flow is linearly stable, so a sufficiently large perturbation is necessary to trigger transition to turbulence.
In this Section, we show that our ESNs can be used to predict transition to turbulence.

To characterize this transition statistically, it is convenient to introduce the laminarization probability $P_{\text{lam}}(E)$, which is the probability that a random perturbation to the laminar flow decays as a function of its kinetic energy $E$ \cite{Pershin2020}.
The laminarization probability is related to the relative volume of the basin of attraction of the laminar flow and is therefore related to the notion of basin stability \cite{Menck2013}.

\begin{figure}
\includegraphics[width=0.49\textwidth]{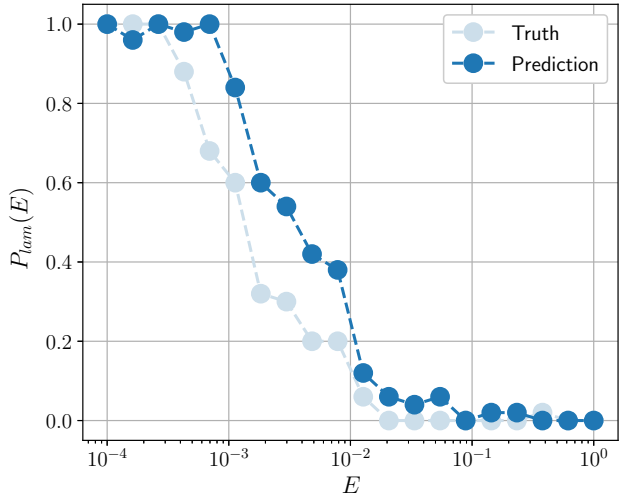}
\caption{\label{fig:p_lam} Laminarization probability as a function of the kinetic energy of random perturbations plotted for the Moehlis--Faisst--Eckhardt model (light blue) and the Echo State Network (bright blue) at $Re = 500$.}
\end{figure}

We compute the laminarization probability for $20$ different values of the kinetic energy of perturbations evenly spaced in the logarithmic scale: $E_1 = 10^{-4}, \dots, E_{20} = 1$.
For each energy level $E_j$, we generate $50$ random perturbations by drawing $a_k(0)$, $k = 1, \dots, 9$, from the uniform distribution with support $[-1; 1]$ and scaling them such that the kinetic energy of perturbations is equal to $E_j$.
We then time-integrate the Moehlis--Faisst--Eckhardt model starting from each of the generated perturbations for $300$ time units and record transition to turbulence if the flow kinetic energy becomes smaller than $10$ within this time window.
The laminarization probability $P_{\text{lam}}(E_j)$ is then approximated as the fraction of those random perturbations which do not lead to transition to turbulence.
We used the same procedure to estimate the laminarization probability using the ESN trained on the turbulent state, except that each perturbation is time-advanced for $10$ time-steps using the Moehlis--Faisst--Eckhardt model to provide sufficient data for synchronization.

The resulting dependence of $P_{\text{lam}}(E)$ on the perturbation kinetic energy $E$ for $Re = 500$ is shown in figure \ref{fig:p_lam} for both the truth and prediction.
The laminarization probability of the original model is almost monotonically decreasing with $E$.
It tends to $1$ for small perturbation energies (the laminar flow is linearly stable) and we found that $P_{\text{lam}}(E) = 0$ for $E \gtrsim 2 \times 10^{-2}$, indicating that all the perturbations beyond this energy trigger transition to turbulence.
The ESN prediction exhibits the same trend and compares well qualitatively with the true $P_{\text{lam}}(E)$, showing that ESNs are capable of learning the statistical boundaries of the basin of attraction of the laminar flow.
It is, in fact, remarkable that the ESN can successfully estimate the threshold for laminar-to-turbulent transition \textit{despite having only been trained on fully turbulent timeseries.}
Despite the qualitatively striking prediction, the ESN does somewhat overestimate the laminarization probability for intermediate perturbation energies, thereby slightly overestimating the nonlinear stability of the laminar flow.
This is likely related to the fact that the ESN generates the laminar state in the presence of $O(10^{-3})$ noise making it more stable to perturbations of very small amplitudes.
As a result, we observe a systematically increased laminarization probability in the interval $5 \times 10^{-3} \lesssim E \lesssim 10^{-2}$ in figure \ref{fig:p_lam}.

\section*{\label{sec:conclusion}Discussion}

In this work, we have shown that Echo State Networks, a class of Recurrent Neural Networks, are able to capture dynamical behavior qualitatively different from that included in their training dataset.
We demonstrated this on a classical example of flow transition where a fluid flow can display two distinct types of behavior, laminar flow and turbulence: the Moehlis--Faisst--Eckhardt model.
In this problem, the transition from laminar flow to turbulence is a finite-amplitude instability, while the reverse transition is a spontaneous escape from a chaotic saddle.
We computed predictions of these transitions via the use of Echo State Networks trained solely on turbulent dynamics and compared them to the ``truth'', which we obtained by directly time-integrating the Moehlis--Faisst--Eckhardt model.

Our Echo State Networks were able to learn laminar dynamics despite not having seen it during training.
In addition, they are capable of successfully reproducing the statistical properties of both types of transition.
Finally, we demonstrated that Echo State Networks can successfully act as generators of early warning signals of transitions by providing the probability of turbulent-to-laminar transition conditioned on the time before transition. 

\begin{figure}
\includegraphics[width=0.49\textwidth]{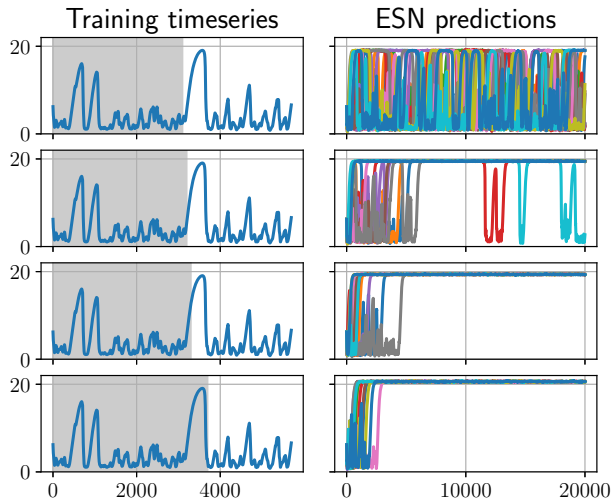}
\caption{\label{fig:learning_laminar_state_timeseries} Predictions made by Echo State Networks (right) each of which was trained using the corresponding shadowed part of the time series shown on the left. Trajectories of different colours on the right were obtained using random initial conditions.}
\end{figure}

This success may be related to the Echo State Network Approximation Theorem recently proved for a one-dimensional observable of the true dynamical system and in the absence of noise \cite{Hart2020}.
It states that, under some mild conditions, for a sufficiently large reservoir and structurally stable true dynamical system, there exists such a matrix $\boldsymbol{W}_{out}$ that the dynamical system defined by the resulting Echo State Network is topologically conjugate to the true dynamical system.
A crucial consequence of this theorem is that an Echo State Network is also expected to embed attractors of the true system.
It however does not provide us with particular rules for building the matrices $\boldsymbol{W}$, $\boldsymbol{W}_{in}$ and $\boldsymbol{W}_{out}$ guaranteeing that a given attractor will be embedded into the manifold generated by an Echo State Network.
We believe that the training time series plays a crucial role in this process.
We found that Echo State Networks failed to learn the laminar dynamics and, as a consequence, were not able to produce any transitions when the training time series did not include at least one large-amplitude excursion pulling the flow relatively close to the laminar state.
This fact is demonstrated in figure \ref{fig:learning_laminar_state_timeseries} where we show predictions made by Echo State Networks trained on four time series.
The first one (top plots) does not cover the large excursion at $t \approx 3500$ whatsoever which results in an Echo State Network being unable to generate laminar dynamics and turbulent-to-laminar transition.
The second one covers only a small piece of this excursion which is enough for an Echo State Network to reproduce laminar dynamics, but in an unstable fashion.
Finally, the third and fourth time series cover a sufficient part of the excursion in order for Echo State Networks to generate stable laminar flows.
This means that Echo State Networks only require training time series to include some dynamics in the vicinity of another dynamical regime without actually visiting it.

We believe that our results provide strong evidence that Echo State Networks can be used for data-driven discovery of new dynamical regimes, early warning of transitions between different dynamical modes and prediction of reversals from transitions.
They have a potential to be useful in a range of applications involving complex nonlinear systems characterized by abrupt transitions between dynamical regimes \cite{Scheffer2009, Ditlevsen2010, Lenton2011, Scheffer2012, Ashwin2012, Kefi2014}.

\acknow{AP is grateful for the support of Professor Timothy Palmer and the European Research Council grant ITHACA (Grant agreement No. 741112) at the University of Oxford.}

\showacknow{} 

\bibliography{2021_rc}

\providecommand{\noopsort}[1]{}\providecommand{\singleletter}[1]{#1}%
\begin{thebibliography}{10}

\bibitem{Xu2021}
K {Xu}, et~al., How neural networks extrapolate: From feedforward to graph
  neural networks in {\em ICLR 2021: The Ninth International Conference on
  Learning Representations}.
\newblock (2021).

\bibitem{Brunton2020}
SL Brunton, BR Noack, P Koumoutsakos, Machine learning for fluid mechanics.
\newblock {\em\protect\JournalTitle{Annual Review of Fluid Mechanics}}
  \textbf{52}, 477--508 (2020).

\bibitem{Goodfellow2016}
I Goodfellow, Y Bengio, A Courville, {\em Deep learning}.
\newblock (MIT press), (2016).

\bibitem{Jaeger2004}
H Jaeger, H Haas, Harnessing nonlinearity: Predicting chaotic systems and
  saving energy in wireless communication.
\newblock {\em\protect\JournalTitle{science}} \textbf{304}, 78--80 (2004).

\bibitem{Lukovsevivcius2009}
M Luko{\v{s}}evi{\v{c}}ius, H Jaeger, Reservoir computing approaches to
  recurrent neural network training.
\newblock {\em\protect\JournalTitle{Computer Science Review}} \textbf{3},
  127--149 (2009).

\bibitem{Vlachas2020}
PR Vlachas, et~al., Backpropagation algorithms and reservoir computing in
  recurrent neural networks for the forecasting of complex spatiotemporal
  dynamics.
\newblock {\em\protect\JournalTitle{Neural Networks}} \textbf{126}, 191--217
  (2020).

\bibitem{Pathak2017}
J Pathak, Z Lu, BR Hunt, M Girvan, E Ott, Using machine learning to replicate
  chaotic attractors and calculate lyapunov exponents from data.
\newblock {\em\protect\JournalTitle{Chaos: An Interdisciplinary Journal of
  Nonlinear Science}} \textbf{27}, 121102 (2017).

\bibitem{Chen2020}
X Chen, et~al., Mapping topological characteristics of dynamical systems into
  neural networks: A reservoir computing approach.
\newblock {\em\protect\JournalTitle{Physical Review E}} \textbf{102}, 033314
  (2020).

\bibitem{Doan2021}
NAK Doan, W Polifke, L Magri, Short- and long-term predictions of chaotic flows
  and extreme events: a physics-constrained reservoir computing approach.
\newblock {\em\protect\JournalTitle{Proc. R. Soc. A}} \textbf{477}, 20210135
  (2021).

\bibitem{Pathak2018}
J Pathak, et~al., Hybrid forecasting of chaotic processes: Using machine
  learning in conjunction with a knowledge-based model.
\newblock {\em\protect\JournalTitle{Chaos: An Interdisciplinary Journal of
  Nonlinear Science}} \textbf{28}, 041101 (2018).

\bibitem{Pathak2018modelfree}
J Pathak, B Hunt, M Girvan, Z Lu, E Ott, Model-free prediction of large
  spatiotemporally chaotic systems from data: A reservoir computing approach.
\newblock {\em\protect\JournalTitle{Phys. Rev. Lett.}} \textbf{120}, 024102
  (2018).

\bibitem{Pandey2020}
S Pandey, J Schumacher, Reservoir computing model of two-dimensional turbulent
  convection.
\newblock {\em\protect\JournalTitle{Physical Review Fluids}} \textbf{5}, 113506
  (2020).

\bibitem{Heyder2021}
F Heyder, J Schumacher, Echo state network for two-dimensional turbulent moist
  rayleigh-b{\'e}nard convection.
\newblock {\em\protect\JournalTitle{Physical Review E}} \textbf{103}, 053107
  (2021).

\bibitem{Haluszczynski2019}
A Haluszczynski, C R{\"a}th, Good and bad predictions: Assessing and improving
  the replication of chaotic attractors by means of reservoir computing.
\newblock {\em\protect\JournalTitle{Chaos: An Interdisciplinary Journal of
  Nonlinear Science}} \textbf{29}, 103143 (2019).

\bibitem{Chattopadhyay2020}
A Chattopadhyay, P Hassanzadeh, D Subramanian, Data-driven predictions of a
  multiscale lorenz 96 chaotic system using machine-learning methods: reservoir
  computing, artificial neural network, and long short-term memory network.
\newblock {\em\protect\JournalTitle{Nonlinear Processes in Geophysics}}
  \textbf{27}, 373--389 (2020).

\bibitem{Barkley2016}
D Barkley, Theoretical perspective on the route to turbulence in a pipe.
\newblock {\em\protect\JournalTitle{J. Fluid Mech.}} \textbf{803}, P1 (2016).

\bibitem{Lucarini2019}
V Lucarini, T B{\'o}dai, Transitions across melancholia states in a climate
  model: Reconciling the deterministic and stochastic points of view.
\newblock {\em\protect\JournalTitle{Physical review letters}} \textbf{122},
  158701 (2019).

\bibitem{Menck2013}
PJ Menck, J Heitzig, N Marwan, J Kurths, How basin stability complements the
  linear-stability paradigm.
\newblock {\em\protect\JournalTitle{Nature physics}} \textbf{9}, 89--92 (2013).

\bibitem{Petrelis2009}
F P{\'e}tr{\'e}lis, S Fauve, E Dormy, JP Valet, Simple mechanism for reversals
  of earth’s magnetic field.
\newblock {\em\protect\JournalTitle{Physical review letters}} \textbf{102},
  144503 (2009).

\bibitem{Tobias2021}
S Tobias, The turbulent dynamo.
\newblock {\em\protect\JournalTitle{Journal of Fluid Mechanics}} \textbf{912},
  P1 (2021).

\bibitem{Moehlis2004}
J Moehlis, H Faisst, B Eckhardt, A low-dimensional model for turbulent shear
  flows.
\newblock {\em\protect\JournalTitle{New Journal of Physics}} \textbf{6}, 56
  (2004).

\bibitem{Waleffe1997}
F Waleffe, On a self-sustaining process in shear flows.
\newblock {\em\protect\JournalTitle{Physics of Fluids}} \textbf{9}, 883--900
  (1997).

\bibitem{Moehlis2005}
J Moehlis, H Faisst, B Eckhardt, Periodic orbits and chaotic sets in a
  low-dimensional model for shear flows.
\newblock {\em\protect\JournalTitle{SIAM Journal on Applied Dynamical Systems}}
  \textbf{4}, 352--376 (2005).

\bibitem{Xiao2021}
R Xiao, LW Kong, ZK Sun, YC Lai, Predicting amplitude death with machine
  learning.
\newblock {\em\protect\JournalTitle{Physical Review E}} \textbf{104}, 014205
  (2021).

\bibitem{Avila2011}
K Avila, et~al., The onset of turbulence in pipe flow.
\newblock {\em\protect\JournalTitle{Science}} \textbf{333}, 192--196 (2011).

\bibitem{Scheffer2009}
M Scheffer, et~al., Early-warning signals for critical transitions.
\newblock {\em\protect\JournalTitle{Nature}} \textbf{461}, 53--59 (2009).

\bibitem{Pershin2020}
A Pershin, C Beaume, SM Tobias, A probabilistic protocol for the assessment of
  transition and control.
\newblock {\em\protect\JournalTitle{Journal of Fluid Mechanics}} \textbf{895}
  (2020).

\bibitem{Lenton2011}
TM Lenton, Early warning of climate tipping points.
\newblock {\em\protect\JournalTitle{Nature climate change}} \textbf{1},
  201--209 (2011).

\bibitem{Ashwin2012}
P Ashwin, S Wieczorek, R Vitolo, P Cox, Tipping points in open systems:
  bifurcation, noise-induced and rate-dependent examples in the climate system.
\newblock {\em\protect\JournalTitle{Philosophical Transactions of the Royal
  Society A: Mathematical, Physical and Engineering Sciences}} \textbf{370},
  1166--1184 (2012).

\bibitem{Ditlevsen2010}
PD Ditlevsen, SJ Johnsen, Tipping points: Early warning and wishful thinking.
\newblock {\em\protect\JournalTitle{Geophysical Research Letters}} \textbf{37}
  (2010).

\bibitem{Kefi2014}
S Kefi, et~al., Early warning signals of ecological transitions: methods for
  spatial patterns.
\newblock {\em\protect\JournalTitle{PloS one}} \textbf{9}, e92097 (2014).

\bibitem{Scheffer2012}
M Scheffer, et~al., Anticipating critical transitions.
\newblock {\em\protect\JournalTitle{science}} \textbf{338}, 344--348 (2012).

\bibitem{Hart2020}
A Hart, J Hook, J Dawes, Embedding and approximation theorems for echo state
  networks.
\newblock {\em\protect\JournalTitle{Neural Networks}} \textbf{128}, 234--247
  (2020).

\end{thebibliography}

\end{document}